# Modeling the dynamics of domain specific terminology in diachronic corpora

Gerhard Heyer[1], Cathleen Kantner[2], Andreas Niekler[1], Max Overbeck[2], Gregor Wiedemann[1]

[1]Leipzig University, Natural Language Processing Group, Computer Science Department
`heyer|aniekler|wiedemann@informatik.uni-leipzig.de`
[2]Stuttgart University, Chair for International Relations, Social Sciences Institute
`cathleen.kantner|maximilian.overbeck@sowi.uni-stuttgart.de`

**Abstract:** In terminology work, natural language processing, and digital humanities, several studies address the analysis of variations in context and meaning of terms in order to detect semantic change and the evolution of terms. We distinguish three different approaches to describe contextual variations: methods based on the analysis of patterns and linguistic clues, methods exploring the latent semantic space of single words, and methods for the analysis of topic membership. The paper presents the notion of *context volatility* as a new measure for detecting semantic change and applies it to key term extraction in a political science case study. The measure quantifies the dynamics of a term's contextual variation within a diachronic corpus to identify periods of time that are characterised by intense controversial debates or substantial semantic transformations.

## 1    Introduction

While the classical theory of terminology presupposes that key terms reflect objective, clear-cut concepts within static conceptual structures (Wüster 1979), recent advances in terminology work have highlighted the dynamics of terms in diachronic text corpora and propose explanations for the change and development of terms (S. Fernández-Silva et. al. 2011, Picton 2011). The methods for key term extraction in computational linguistics and terminology engineering can roughly be divided into *frequentist* and *Bayesian* approaches. On the one hand, focusing on the frequency of terms, statistical tests such as log-likelihood-ratio can be employed to compare expected with observed term frequencies using reference corpora (Archer 2008). To detect changes in a term's usage, it is also common to observe a term's context and evaluate how it may change over time (Lenci 2008). By this approach, contextual variations can be measured using a bag of words document model and thresholds based on a tf/idf comparison of text stream segments (e.g. Kumaran and Allan, 2004). On the other hand, assuming a *Bayesian model* of topic and term distribution in documents, one can also use co-occurrence patterns and their local distribution in time to detect changing topics over time (Wang & McCallum 2006).

In most diachronic corpora, however, the patterns for the emergence of new terms, or contextual changes of existing terms, cannot be described just by reference to frequency or topic clusters (S. Fernández-Silva et. al. 2011). Rather, they are the result of a number of factors such as *centrality*, i.e. the use of terms and concepts to convey a change in the domain where the terms "all belong to a common topic in the domain and indicate an evolution in this topic" (Picton 2011, p. 147). Often, the increase or decrease of occurrences of terms in a domain is not related to novelty, but to the centrality/disappearance of a topic in the domain of application because of scientific or public discussion (ibid.).

In order to better describe and track controversial discussions reflected in diachronic corpora, we would like to introduce the notion of *context volatility*. Assuming a distributional model of meaning (Turney & Pantel 2010), we consider a term's global context (see below) as a second dimension for analyzing its salience and temporal extension in addition to term frequency. Changes over time in the global context of a term thus indicate a change of usage. Our novel approach differs from previous ones in the spirit of distributional semantics in important aspects: for us the *rate of change* is indicative of how much the "opinion stakeholders" agree, or disagree, on the meaning of a term. Fixing the usage of a term within a community of speakers seems in some ways similar to fixing the price of a stock at a stock market. Reversely, the analysis of the volatility of a term's global context can be employed to detect controversial or changing topics. In the following, we will first review related work on contextual variation of terms, and then explain the basic notions and assumptions of our approach. Finally, we will present first experimental results from a case study carried out in political science.

## 2     Context Change of Terms – Related Work

In terminology work, natural language processing, and digital humanities, several studies address the analysis of variation in context of terms in order to detect semantic change and the evolution of terms. Three different approaches to describe contextual variations can be distinguished: (1) methods based on the analysis of patterns and linguistic clues to explain term variations, (2) methods that explore the latent semantic space of single words, and (3) methods for the analysis of topic membership.

(1) Most studies in the area of terminology focus on particular terms, and look for linguistic clues and different patterns of variation in their usage to better understand the dynamics of terms such as Fernández-Silva, Freixa, and Cabré (2011) or Picton (2011). These studies take a particular term as starting point and inspect its neighbouring context to classify, analyse and predict changes of usage. In contrast, our approach takes a whole corpus as starting point, and aims at detecting terms that exhibit a high rate of contextual variation for some time.

(2) In NLP and digital humanities, distributional properties of text have been used to study the dynamics of terms in diachronic texts. Jatowt and Duh (2014) use latent semantics of words in order to create representations of a term's evolution. Hilpert (2011) proposed a similar method, which uses multidimensional scaling to find latent semantic structures, and compare them for different periods. These approaches try to model semantic change over time by setting a certain time period as reference point and comparing the latent semantic space to that reference over time. Terms can thus be compared with respect to their semantic distance or similarity over time. Again, our approach differs from these because we do not start with a fixed set of terms to study and trace their evolution, but rather we want to detect terms in a collection of documents that may be indicative of semantic change.

(3) Assuming a Bayesian approach, topic modeling is another method to analyse the usage of terms and their embeddedness within topics over time (Rohrdantz et al. 2011; Rohrdantz et al. 2012). These studies identify terms, which have changed in usage and context, and show that this change can be quantified by the probability of a term's membership in a topic cluster within the topic model used. Approaches like the one of Blei and Lafferty (2006) model the dynamics of a term's topic membership directly and allow the model to slightly change its co-occurrence structure over time. Zang et al. (2010) modify hierarchical Dirichlet processes to measure the changing share of salient topics over time, and thus help to identify topics and terms that for are very prominent for some time. Jähnichen (2015) has extended this approach to identify topics that for some period of time contain rapidly changing terms, and thus can be considered to be indicative of conceptual changes. However, topic model based approaches always require an interpretation of the topics and their context. In effect, the analysis of a term's change is always relative to the interpretation of the global topic cluster, and strongly depends on it. Topic models only generate a *macro view* on document collections. In order to identify contextual variations, we also need to look at the key terms that drive the changes at the *micro level*. Often these hot-button words fan the flames of a debate.

In sum, while related work on the dynamics of terms usually starts with a reference (like pre-selected terms, some pre-defined latent semantics structures, or given topic structures), we aim at automatically identifying terms that exhibit a high degree of contextual variation in a diachronic corpus. The typological category of *centrality* as introduced by Picton (2011) tries to capture the observation that central terms simultaneously appear or disappear in a corpus when the key assumptions, or consensus, amongst the stakeholders of a domain change. The measure of *context volatility* is intended to support exploratory search for such central terms in diachronic corpora, in particular, if we want to identify periods of time that are characterised by substantial semantic transformation. However, we do not claim that our measure quantifies meaning change or semantic change, the measure quantifies the dynamics of a term's contextual information within a diachronic corpus.

## 3   Context Volatility - Intuition

Our focus for identifying context changes is on the retrieval of what *authors* consider "worth writing about" (for whatever reason). Any topic "worth writing about" represents some author's point of view (at some point of time). On some topics there may be agreement, others may be contested – and this can change over time. "Hot-button" topics are highly controversial topics with a clear-cut distinction between proponents and opponents.

We observed that for competing opinion stakeholders, the linguistic context of key terms is *different*. For example, with the exception of the controversial term "nuclear power" and some stop-words, there is no overlap between the controversial positions on nuclear power based on excerpts from internet fora summarized below (table 1).

| *Pro* nuclear power | *Contra* nuclear power |
|---|---|
| Nuclear power is a very efficient source of energy. It is also abundant, unlike fossil fuels (coal and oil). | Nuclear power plants are hard to control. Like in Fukushima 2011, a steam buildup in a nuclear reactor in Chornobyl, Ukraine, caused an explosion that released tons of radiation into contact with people and animals. The radiation released from nuclear fission is harmful to living organisms. |

*Table 1: Controversial positions on "nuclear power"*

When dealing with real-life time-stamped data spanning long periods of time (e.g. newspaper texts, patent applications, or scientific publications), we observed, moreover, that the *global context* of terms does not need to be static, but may radically change. The global context of a term – we assume – consists of all its statistically significant co-occurrences within a corpus, where we measure significance using the log-likelihood ratio (Heyer et al., 2008).[1] To give an example, consider the changes in the global context of the German term "*Kredit*" (credit/loan) in the digital edition of the German weekly newspaper DIE ZEIT. Co-occurrence statistics computed on a yearly basis and visualized as context-networks display almost complete changes of the semantic context (see figure 1 for the graphs for the years 2005, 2007 and 2009).

---

[1] A term's set of co-occurrences is computed on the basis of the term's joint appearance with its co-occurring terms within a predefined text window taking an appropriate measure for statistically significant co-occurrence. The global context can also be displayed as a graph which contains the term and its context terms as nodes where the edges have a weight according to the significance value of the joint appearance of the terms.

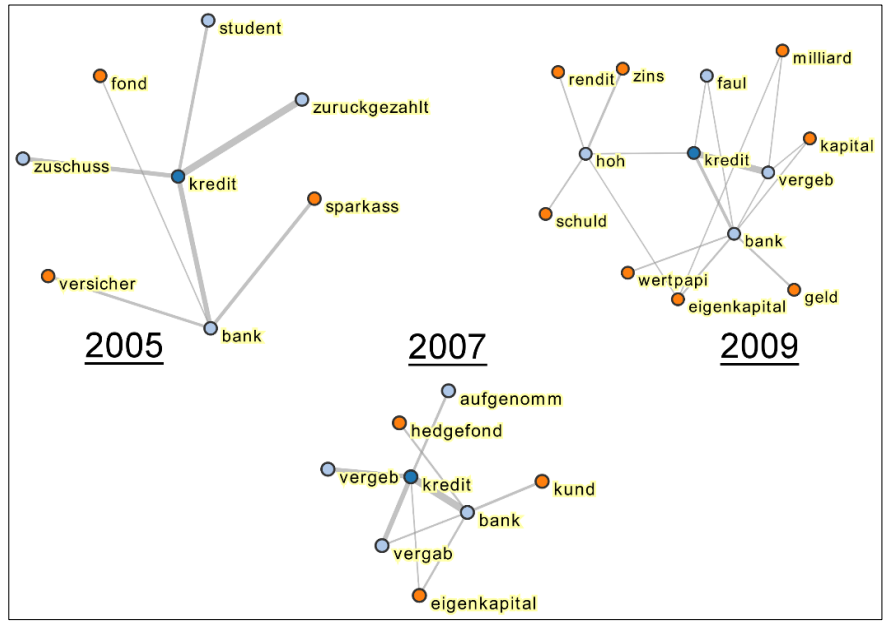

*Figure 1: Changes in the global context of the German term "Kredit" (credit/loan)*

While in 2005 the main usage apparently covered references in the context of student loans, in 2007 there is already a mention of net assets *(Eigenkapital)* in connection with credits granted by banks. Finally, in 2009, the modifier *hoch* (*high*) is linked to *Zins* (interest rates), *Rendite* (income return) and *Schuld* (debt). Furthermore, we see the evaluator *faul* (foul) linked to the word *Kredit* (loan). Quite obviously, the risks taken by banks granting bad credits was something worth reporting on, and by doing so, the global context of "*Kredit*" has changed substantially so that the link between *faul* and *Kredit* became almost collocational. Following this approach, a new multi-term expression can be viewed as a new term referring to the way banks were handling credits in 2009.

## 4    Context Volatility – Definition

The basis of our analysis is a set of time stamped text corpora, e.g. all editions of a digital weekly newspaper between January 2005 and December 2010 which is our test case in this paper. Our measure of the contextual changes is the mean *volatility* in the co-occurrence ranks of a term. It is inspired by the widely used risk measure in econometrics and finance[2], and based on the ranking of significant co-occurrences in a defined time slice. A time slice is a set of documents belonging to a consecutive time span. The corpus is divided into time spans allowing, however, for various options from years, months, weeks, days or even hours to minutes. The example in this paper was created using months as the time spans of choice. Informally, we compute a term's change of context by averaging the changes in the *ranks* of its co-occurrences for a defined number of time slices. This can be conducted in a variety of ways. We considered all time slices in order to define a global measure of the dynamics of a term's context, e.g. the changes of its distributional semantics. Moreover, we also build the measure for a window of time slices for each term to produce a time series of a term's context change. *Context volatility* is then computed as the *average* of all rank changes of a term's co-occurrences for some period of time as follows:

1. Compute for every word $w$ of the vocabulary $V$ and every time slice $t$ (days, weeks, years) in the data of all time slices $T$ the set of co-occurrences, e.g. a term-term matrix $C_t$ with co-occurrence weights for every time slice. The matrix has the dimension $V \times V$.[3]

2. Compute for every word the rank for every concurrent word for every time slice as a matrix $R_{V,T}$ where the rows represent the ranks of all co-occurent words of $w$ throughout the time slices. This matrix has the dimension $V \times T$ and is produced for every word in $V$.

3. Compute the *context volatility* of a word for a given history h in the time slices $T$ by computing the difference between the 3$^{rd}$ and the 1$^{st}$ quartile of all ranks that the co-occurents of word w take for all time slices in h, e.g. the interquartile range (IQR) of a row in $R_{w,T}$ where we limit the row to $t$ elements of $h$. The result is again a matrix $CV_{w,T}$ where each row contains the IQR at a time slice $t$ for a given history $h$.

---

[2] Yet, it is calculated differently and not based on widely used gain/loss measures. For an overview of miscellaneous approaches to volatility see Taylor (2007).

[3] The weights can be set by significance measures like Log-likelihood, Dice, Mutual Information or a significance test based on the Poisson distribution. For this paper we used the log-likelihood significance measure.

4. Compute the global *context volatility* for a word w by averaging the columns, e.g. all co-occurents in $CV_{w,T}$, to compute the mean of all standard deviations in the rank changes. The result is a vector $S_w$ which represents the quantity of context change as defined by the *context volatility* w.r.t the defined sequence of back-looking windows. If we define the back-looking history as the set of all time slices within the data, we get a single constant. If $h$ is a window shorter than $T$ we get a time series of quantified context changes for that term with the length $T-h$. In summary, we can define the final calculation of the volatility for h or T as

$$CV_{w,T} = \frac{1}{C_{w,T}} \sum_i IQR\left(Rank(C_{w,i}, T)\right)$$

Here $C_{w,T}$ is the number of all co-occurences of w in $T$. $C_{w,i}$ is the ith co-occurence of *w* in *T*. *Rank* represents a set of all ranks $C_{w,i}$ holds within *T*, and IQR is the interquartile range of those ranks.

As this computation is complex (at least $O|n^2 * t|$ with n the size of the vocabulary and t the number of time slices), we improved the runtime of our algorithm by considering only the overall most significant co-occurrences (filtering out stop-words and pruning words with a document frequency < 3). We also used parallel computations to speed up the process. We parallelized the computation of the matrices $C_t$ since they are totally independent from each other. Furthermore, we parallelized the computation of $R_{w,T}$, $CV_{w,T}$ and $S_w$ to compute their values for every term separately. This way the whole process is scalable w.r.t $T$ and V.

## 5 Use case – Issue Analysis in Political Science

The measure of context-volatility enables us to explore large amounts of documents and to identify periods of substantial semantic change. This opens fruitful ways for the identification of so-called "issues" in public political communication. A political issue is "a controversial social problem, which constitutes a broader topical structure, encompassing several events as belonging together" (Kantner 2015, p. 40). Social problems are real-world matters involving a certain vocabulary. However, events, actors, opinions, cultural and technical features change over time. This results in a dilemma, especially when we want to identify issues over longer historical periods: On the one hand, we want to identify terms that characterize issues as some kind of generic social problem that at some points in time provoke intense and controversial discussions, and for which at different times different solutions have been proposed. On the other hand, we also want to identify those periods of time where the issue is being fed by new conflicts, contested, and redefined and thus undergoes semantic transformation. Therefore, we are interested in, both, terms that describe issues in general irrespective of contextual variation and semantic change, and at the same time exactly those terms that mark particular periods of crisis and semantic change within the issue.

In order to deal with that dilemma, we proceeded in two steps combining topic modeling with context-volatility analysis. Our use case is based on 397,729 articles from altogether 3,841 editions of the German weekly newspaper DIE ZEIT covering the period from 1946 – 2011.[4] One central problem with standard topic-models is that they generate topics that are not intuitive and that they involve largely named entities such as people, places, and events. To compute the generic political issues for this document collection we, therefore, computed in a first step 30 topics based on the Latent Dirichlet Allocation Model (LDA) (Blei et al. 2003) after deleting all named entities such as names of people, places, and events. Since issues are defined as broader social problems, named entities referring to those people, places, and events, characterize an issue during a short time span. To delete the event-bias and to catch only the properties of the issues in general, the topic model was created without named entities.

Thirty topical fields could be distinguished. Among them, one topic relates to *"financial and economic policies"* (fig. 2). For the remainder of this paper we will focus on this relation as name for the topic represented by the 30 most probable terms inferred by the LDA model.

---

[4] The data were retrieved from DTA corpus. The preprocessing of the text sources includes the following steps: sentence segmentation, tokenization, named entity recognition, multi-word unit identification, stop-word deletion, lower case transformation and lemmatization. The resulting term vectors for each sentence where used to create a sentence-term matrix for annual time slices. Those matrices where pruned to delete high frequent and low frequent words from the process. We used relative pruning and excluded vocabulary which is found in more than 99% and in less than 1% of the documents.

> dollar, milliarde, jahr, prozent, geld, million, gewinn, zins, kredit, markt, fond, pfund, geschäft, kasse, bank, unternehmen, verlust, währung, investor, kunde, umsatz, anteil, konzern, schuld, investition, gold, verkauf, monat, versicherung, kauf

*Figure 2: Words representing the topic "financial and economic policies"*

When looking at the temporal salience of *financial and economic policies*, we clearly see changing phases of activity, e.g. high peaks in the early seventies relating to discussions of currency parities, or in 2008/2009 relating to the last financial crisis (fig. 3). The longitudinal data was produced by counting those documents within the corpus, which contain the context, e.g. topic (see fig. 2), from our inferred LDA model at a minimum of 30%.

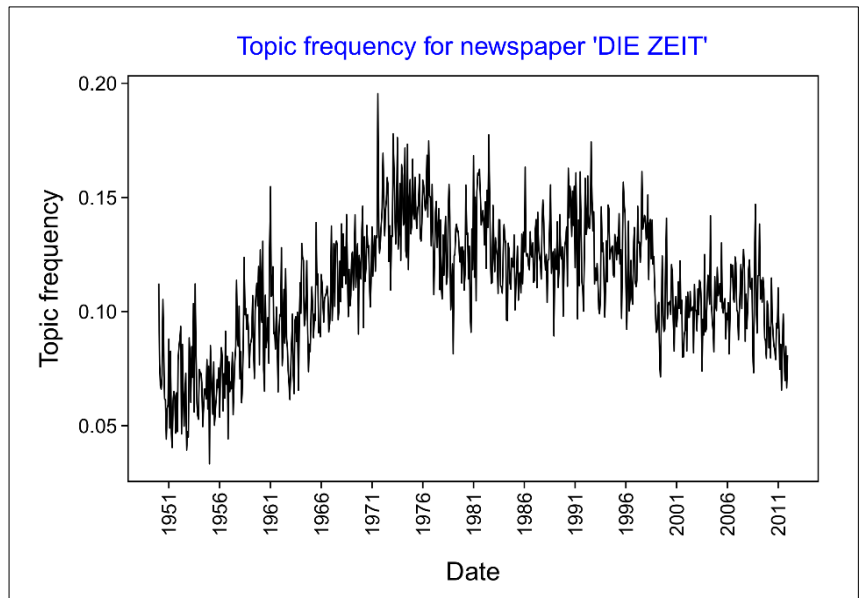

*Figure 3: Temporal salience of topic, normalized, monthly basis*

In order to identify issues in the technical sense, we then identified key terms within that topic that not only have a high relative frequency, or tf/idf-value[5], but can also be considered to fuel controversial discussion. Thus, by looking for key terms in controversies, and by assuming that the context of these terms is rapidly changing, the measure of *context volatility* is a natural choice. In our case study, we wanted to test whether our *context volatility* measure is able to recognize the last financial crisis in 2008/2009. In order to do so, we applied the measure on a suitable sub-corpus of the whole data for one topic (*financial and economic po*licies) and the years 2005 to 2010. This time we included the named entities again and, of course, we were pointed to some of these named entities that are characteristic for that period of time, and that describe the key actors of the crisis such as *Lehman Brothers,* or *Goldman-Sachs*. However, we also found terms like *Kredit* (loan*), Banken* (banks)*, Fonds* (fonds) and *Schulden* (debts)*,* that are constitutive of the general topic (fig. 4). Again, we constructed the global terms from the LDA model with the 30 most probable words from the topic. The top volatile terms created by our measure applied for all time slices (in months) of our corpus.

| Important terms in financial and economic policies topic (1946-2011) | Top volatile terms in financial markets sub-corpus (2005-2010) |
|---|---|
| dollar, milliarde, jahr, prozent, geld, million, gewinn, zins, kredit, markt, fond, pfund, geschäft, kasse, bank, unternehmen, verlust, währung, investor, kunde, umsatz, anteil, konzern, schuld, investition, gold, verkauf, monat, versicherung, kauf | dollar, bank, kredite, anleger, unternehmen, geld, banken, schulden, wert, gewinn, umsatz, dresdner, lehman, goldman, zinsen, investieren, merkel, morgan, pfund, währungsfonds, wunder, zentralbank, aktien, estate, fonds |

*Figure 4: Comparison of global terms (topic) and top volatile-terms (context volatility over all time slices) in financial markets sub-corpus*

From a methodological point of view, it is interesting to notice that *context volatility* of these terms highly correlates in time with intense public controversy, but not with the terms relative frequency. In figure 5, the *context volatility* and the relative frequency have been plotted for the terms *Kredit* and *Fond*. Both terms are good examples due to their strong context fluctuation within our exemplary issue. The ranges of values were aligned in order to overlay both longitudinal plots. We set a history *h* for the calculation of the *context volatility* of 6 months. The co-occurrence statistics were calculated for each month, which corresponds to monthly time slices. This means that we calculated a *context volatility* for each word at a time *t* based on the contextual changes from the last 6 month. The figures show that the relative word frequency does not correlate with the *context volatility*. Apparently, the possible change of context, the discursivity, salience, or centrality of a term, cannot fully be reflected by its frequency of usage. Longitudinal *context volatility* signals for terms, which in turn can be used to identify points in time where a semantic, or paradigmatic, change of the meaning of a term might happen. Further interpretations could be that the striking term is discussed from different points of view and *context volatility* thus reflects controversial discussion, or it can even be considered a weak signal for new adjustments within mainstream or established contexts. Of course, we can also calculate the volatility for the whole time span of the corpus highlighting terms, which appear in different contexts more often than other terms (see fig. 4).

---

[5] Tf/idf (term frequency / inverse document frequency) values represent the uniqueness and importance of terms within a document (Manning et. al. 2008).

For social scientists, the use of this measure of *context volatility* is highly profitable. With the growing accessibility of very large, long-time textual corpora, scholars are increasingly interested in (and dependent on) the use of automated textual analysis techniques in order to conduct comparative media studies, or to analyze parliamentary debates or presidential speeches. They want to grasp the salience of specific issues over time and among different countries. They are interested in identifying dominant discourses and frames of interpretation in public debates on issues such as immigration or foreign policy. Last but not least, they want to know which actors or organizations are the ones that are most visible in the media in light of important events such as the current refugee crisis in Europe or during the war in Libya 2011. In this regard, measuring the *context volatility* of terms or topics has pioneering character. Social scientists so far could only come to terms with these questions by measuring the frequency of key terms, term or collocation lists, or topic models. However, by measuring the volatility of co-occurring word contexts, they can now approach a second crucial dimension to determine the salience of an issue: The degree of *contentiousness* of a specific term or topic. Assuming that an issue can be understood as an ongoing flow of communication on matters, which are controversially discussed among different stakeholders, it can be concluded that a topic is not only relevant because it is highly frequent in a given amount of textual data. Hot-button issues might moreover be characterized by high variance of their linguistic contexts. Public stakeholders, due to their different views on the same subject, use different terminologies and try to push their opinion in the public contest of opposing convictions. Thus, as depicted in figure 5, it is important to consider both – frequency as well as *context volatility* – in order to best determine the salience of an issue or term. Otherwise, the importance of those terms that are highly frequent might be overrated while the salience of those (even low frequent) terms that have a high degree of *context volatility* are neglected.

## 6   Conclusion

In this paper, we introduced the notion of *context volatility* as a new measure to identify semantic change of key terms and issues in specialized domains of discourse. Our case study in the field of political science focusing on the analysis of political issues demonstrated the usefulness of this measure. It was possible to identify controversial issues marked by certain key-terms that are in general characteristic for the issue as well as some key-terms highly dependent on particular circumstances and crisis situations– such as *Kredit* or *Fond* for the last financial crisis 2008/09. Yet, the usefulness of the new measure of *context volatility* is, of course, not restricted to this area of application. Because it helps to distinguish clearly between the frequency and contextual usage of terms, it may also be of use in other domains of scientific analysis, the identification of new terms in marketing studies, or technology mining, and terminology extraction in diachronic corpora – especially in cases where rather static standard methods prove to be unable to deal with semantic change.

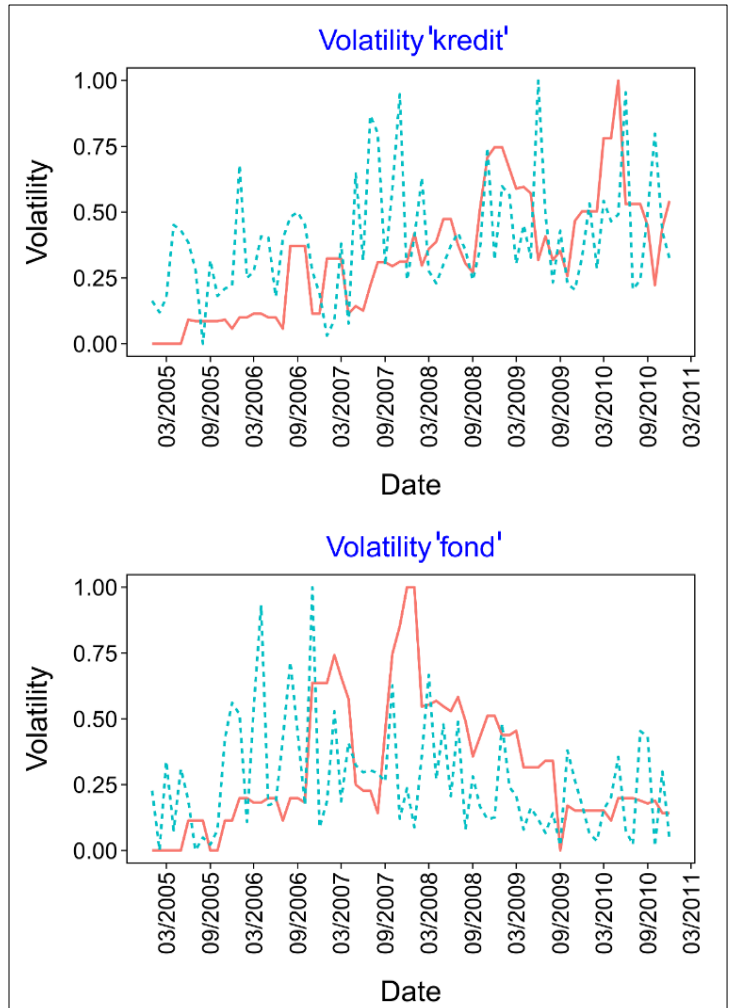

*Figure 5: Relative term frequency (dotted line) and volatility of key terms (solid line)*

## References

Archer, D., (ed.)(2008). What's in a word-list? Investigating word frequency and keyword extraction. Ashgate, Aldershot.

Blei, D. M., Ng, A.Y., and Jordan, M.I. (2003). *Latent dirichlet allocation*. The Journal of Machine Learning Research 3: 993–1022.

Blei, D. M. and Lafferty, J. D. (2006). *Dynamic topic models.* In Proceedings of the 23rd international conference on Machine learning, 113–120.

Harris, Z. (1954). *Distributional Structure*, Word 10 (23): 146-162.

Heyer, G., Quasthoff, U. and Wittig, T. (2008): *Text Mining – Wissensrohstoff Text: Grundlagen, Algorithmen, Beispiele*, Bochum, w3l-Verlag.

Hilpert, M. (2011): *Dynamic Visualizations of Language Change: Motion Charts on the Basis of Bivariate and Multivariate Data from Diachronic Corpora*. International Journal of Corpus Linguistics 16 (4): 435–61.

Jatowt, A. and Duh, K. (2014): *A framework for analyzing semantic change of words across time*. In Proceedings of the 14th ACM/IEEE-CS Joint Conference on Digital Libraries, 229–38.

Jähnichen, P. (2015): Topics over time – *A new approach to dynamic topic models*, Ph.D. Thesis, Leipzig University.

Kantner, C. (2015): *War and Intervention in the Transnational Public Sphere: Problem-Solving and European Identity-Formation*. London, Routledge.

Kumaran, G.; Allan, J. (2004): *Text classification and named entities for new event detection.* In SIGIR '04: Proceedings of the 27th annual international ACM SIGIR conference on Research and development in information retrieval, pages 297–304, New York, NY, USA. ACM.

Lenci, A. (ed.) (2008): *From context to meaning: distributional models of the lexicon in linguistics and cognitive science*, Italian Journal of Linguistics, 20(1): 1-31.

Manning, C. D., Raghavan P. and Schütze H. (2008): *Introduction to Information Retrieval*. New York: Cambridge University Press.

Picton, A. (2011): *Picturing Short-Term Diachronic Phenomena in Specialised Corpora. A Textual Terminology Description of the Dynamics of Knowledge in Space Technologies.* Terminology, 17(1), 134-156.

Fernández-Silva, S., Freixa J. and Cabré, M. T. (2011): *A proposed method for analysing the dynamics of cognition through term variation.* Terminology 17(1). p. 49-73. Amsterdam: John Benjamins.

Taylor, S. (2007): *Asset Price Dynamics, Volatility, and Prediction.* Princeton and Oxford, Princeton University Press.

Turney, P. D. and Pantel, P. (2010): *From Frequency to Meaning.* Vector Space Models for Semantics, in: Journal of Artificial Intelligence Research 37: 141-188.

Rohrdantz, C., Hautli A., Thomas Mayer, Miriam Butt, Daniel A. Keim, und Frans Plank (2011): *Towards Tracking Semantic Change by Visual Analytics*. In Proceedings of the 49th Annual Meeting of the Association for Computational Linguistics: Human Language Technologies: Short Papers - Volume 2, 305–310.

Rohrdantz, C., Niekler, A., Hautli A., Butt M. and Keim, D. A. (2012): *Lexical Semantics and Distribution of Suffixes: A Visual Analysis.* In Proceedings of the EACL 2012 Joint Workshop of LINGVIS & UNCLH, 7–15.

Wang, X. and McCallum, A. (2006): *Topics over time: a non-Markov continuous-time model of topical trends*. In KDD '06: Proceedings of the 12th ACM SIGKDD international conference on Knowledge discovery and data mining, 424–433.

Wüster, E. (1979): *Einführung in die Allgemeine Terminologielehre und Terminologische* Lexikog-raphie. Viena, Springer.

Zhang,J. et. al. (2010): *Evolutionary Hierarchical Dirichlet Processes for Multiple Correlated Time-varying Corpora*. In Proceedings of the 16th ACM SIGKDD international conference on Knowledge discovery and data mining.